\documentclass[11pt]{article}

\usepackage{acl}

\usepackage{times}
\usepackage{latexsym}

\usepackage[T1]{fontenc}

\usepackage[utf8]{inputenc}

\usepackage{microtype}

\usepackage{inconsolata}

\usepackage{graphicx}
\usepackage{booktabs}
\usepackage{multirow}
\usepackage{colortbl}
\usepackage{amsmath,amssymb}
\usepackage{tcolorbox}
\usepackage{placeins}
\definecolor{promptfieldcolor}{HTML}{A12B32}
\newcommand{\promptfield}[1]{\textcolor{promptfieldcolor}{#1}}
\newenvironment{promptbox}{%
    \par\addvspace{8pt}\noindent\begin{minipage}{\linewidth}
    \begin{tcolorbox}[
    colback=black!4,
    colframe=black,
    fontupper=\small\ttfamily,
    before upper={\raggedright\setlength{\parskip}{0.5em}},
    boxrule=0.5pt,
    arc=1.5mm,
    left=7pt,right=7pt,top=6pt,bottom=6pt,
    before skip=0pt,after skip=0pt]
}{\end{tcolorbox}\end{minipage}\par\addvspace{8pt}}

\newcommand{\method}{\textsc{PreView}}
\title{\method: Enhancing GUI Grounding with Preprocessed Views and Cues}

\author{Wenjun Hou \textnormal{and} Kaishuai Xu \\
    Independent Researcher \\
  \texttt{\{houwenjun060, xu.kaishuai\}@gmail.com} \\}

\begin{document}
\maketitle

\begin{abstract}
GUI grounding requires locating small controls in cluttered screenshots and identifying which control a natural-language instruction refers to.
Cropping and enlargement can expose local detail, but the resulting image may still leave ambiguity about the intended target.
We propose \method{}, an auxiliary preprocessor that improves the inputs to an existing grounding model without further training that model.
\method{} prepares two complementary types of input features: task-relevant views for localization and candidate-specific cues for target understanding.
A trained lightweight auxiliary model predicts and refines a center, crops the screenshot into a local view, and enlarges that view for zoomed-in grounding.
It also prepares a visual cue (i.e., a red-circle marker with a location hint) and a textual cue describing the candidate's identity or function.
The frozen grounding model uses the prepared inputs to predict the final click.
Across five benchmarks, \method{} improves the average grounding accuracy of Qwen3-VL-2B, OS-Atlas-4B, and Qwen3-VL-4B without further training these grounding models.
Extensive analyses examine the contributions of views and cues, center refinement, and visual input budgets.
\end{abstract}

\begin{figure}[t]
    \centering
    \includegraphics[width=\columnwidth]{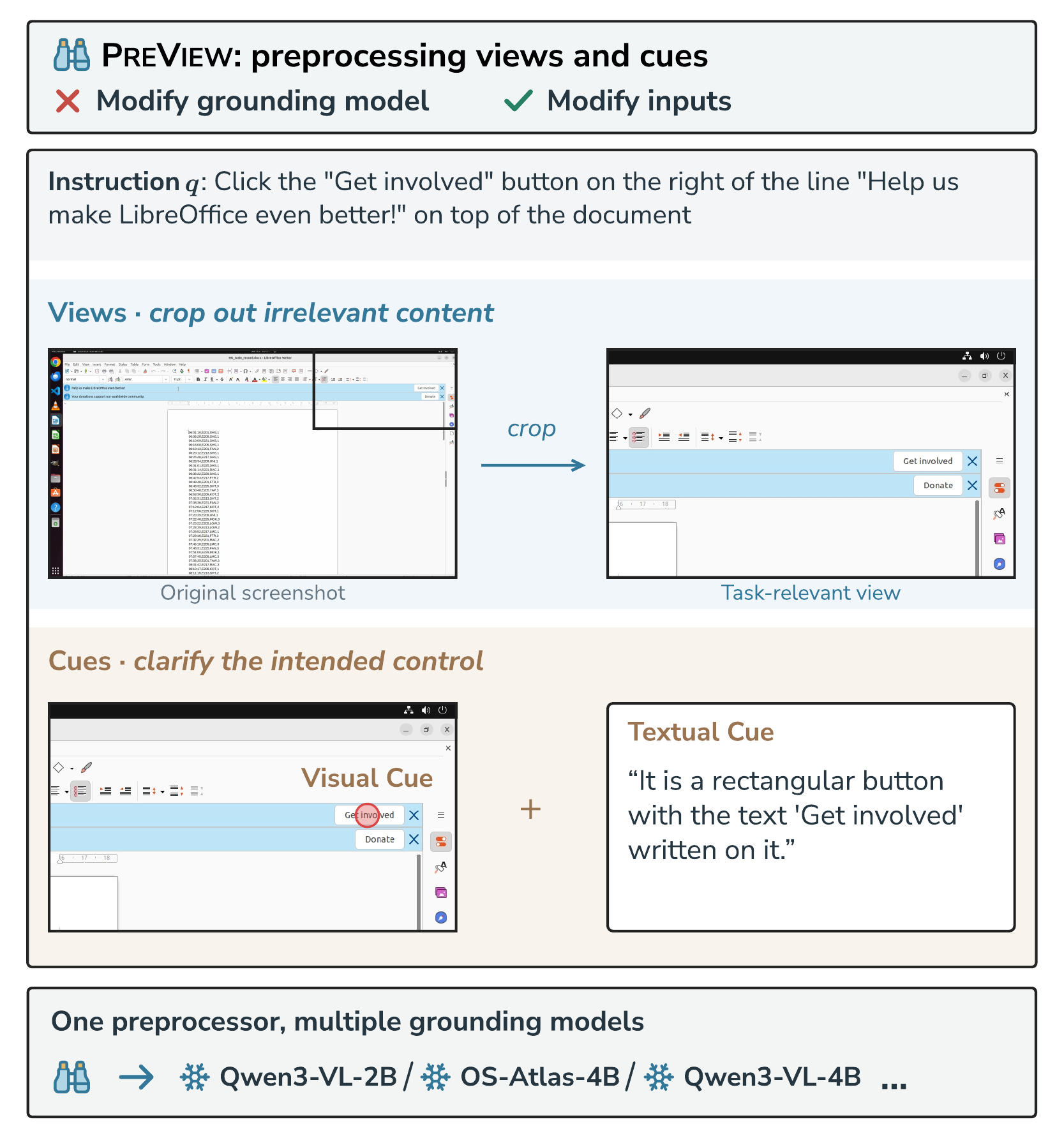}
    \caption{\method{} prepares views and cues at the input level, without modifying the grounding model or accessing its internal states.}
    \label{fig:motivation}
\end{figure}
\section{Introduction}
\label{sec:introduction}

GUI grounding translates a natural-language instruction into the screen location of an actionable interface element.
It connects an agent's intended action to its execution: even a correct plan fails if the agent clicks the wrong control.
Dedicated grounding models have advanced this capability through GUI-specific data and training, including SeeClick, UGround, OS-Atlas, and GroundNext \citep{cheng-etal-2024-seeclick,ICLR2025_4ca0e369,osatlas2024,groundcua2025}.
Professional interfaces remain challenging because they combine high-resolution screenshots, dense layouts, and small interactive elements \citep{screenspotpro2025,uivision2025}.
These settings raise a question beyond the capabilities of the grounder itself: \emph{how should the screenshot and instruction be presented to make the grounding decision easier?}

We consider two intertwined requirements: \emph{localization} and \emph{understanding}.
Localization requires resolving an actionable control's precise position amid competing visual content.
When an entire screenshot is processed under a limited visual budget, small controls may receive insufficient detail.
Understanding requires determining which element satisfies the instruction, including its functional role, its relationship to other elements, and the object to which it belongs.
Consider the instruction to click ``Get involved'' in Figure~\ref{fig:motivation}.
The screenshot contains two stacked notification bars, each with an action button and a close control.
A clearer image can help resolve these controls, but selecting the intended one also requires associating the button with the requested notification.
Conversely, identifying the correct control does not guarantee an accurate click on its small target area.

Existing work addresses these difficulties along several complementary routes.
Training on broader and higher-quality GUI data strengthens the grounder's learned representations \citep{osatlas2024,groundcua2025}.
At inference time, zoom-in methods crop and rescale selected regions to expose local detail, while repeated predictions provide opportunities to revise a localization decision.
ScreenSeekeR uses a planner to guide cascaded visual search \citep{screenspotpro2025}.
ZoomClick iteratively narrows the view around predictions \citep{zoomclick2025}.
UI-Zoomer uses prediction uncertainty to decide whether and how to zoom \citep{uizoomer2026}.
These methods demonstrate that grounding depends on the observation and inference procedure as well as model parameters.
Additional searches or predictions consume inference resources.
Tighter cropping can also discard context needed to distinguish the target from nearby alternatives.

Semantic augmentation provides another source of support.
Screen parsing can expose interactable elements and their functional descriptions, as in OmniParser \citep{omniparser2024}.
Instruction refinement can make an intended target more explicit.
AdaZoom-GUI combines such refinement with conditional zooming \citep{adazoomgui2026}.
These approaches show that visual focusing and semantic clarification can be complementary.
They also motivate a more specific question: \emph{can a learned auxiliary model prepare both task-relevant views and informative cues for an existing grounder?}
Such preprocessing must select useful evidence while limiting two sources of error: removing necessary visual context and supplying misleading semantic descriptions.

We propose \method, an auxiliary model that previews the interface before the downstream grounding decision, without further training the grounding model.
It prepares two complementary types of input features: task-relevant \emph{views}\footnote{Throughout this paper, \emph{crop} denotes the image-selection operation, whereas \emph{view} denotes the image input.} and candidate-specific \emph{cues}.
Given a screenshot and instruction, it predicts and refines a center, then crops the screenshot around that center to produce a local view.
Cropping removes potentially distracting surroundings.
We then zoom in by enlarging the selected view to expose local detail.
The \emph{visual cue} pairs a red-circle marker with a fixed location hint.
When the auxiliary model accepts the candidate, the \emph{textual cue} describes its identity, function, or local relationships.
The frozen grounding model receives the prepared view, the original instruction, and available cues.
Its predicted click is mapped back to the original screenshot.
We jointly train the auxiliary model for center prediction and candidate verification with description generation.

Our contributions are:
\begin{itemize}
    \item We propose \method, an auxiliary preprocessing framework that improves GUI grounding without training or modifying the downstream grounding model.
    \item We introduce two complementary input features: center-guided views that support localization, and candidate-specific visual and textual cues that support target understanding.
    \item Across five benchmarks and three grounding models, we improve reported average accuracy by $1.5$--$13.9$ points, with gains on SSPro, OSW-G, MMB-GUI, and UI-V. Extensive analyses examine feature contributions, refinement, visual budgets, and failures.
\end{itemize}
\begin{figure*}[t]
    \centering
    \includegraphics[width=\textwidth]{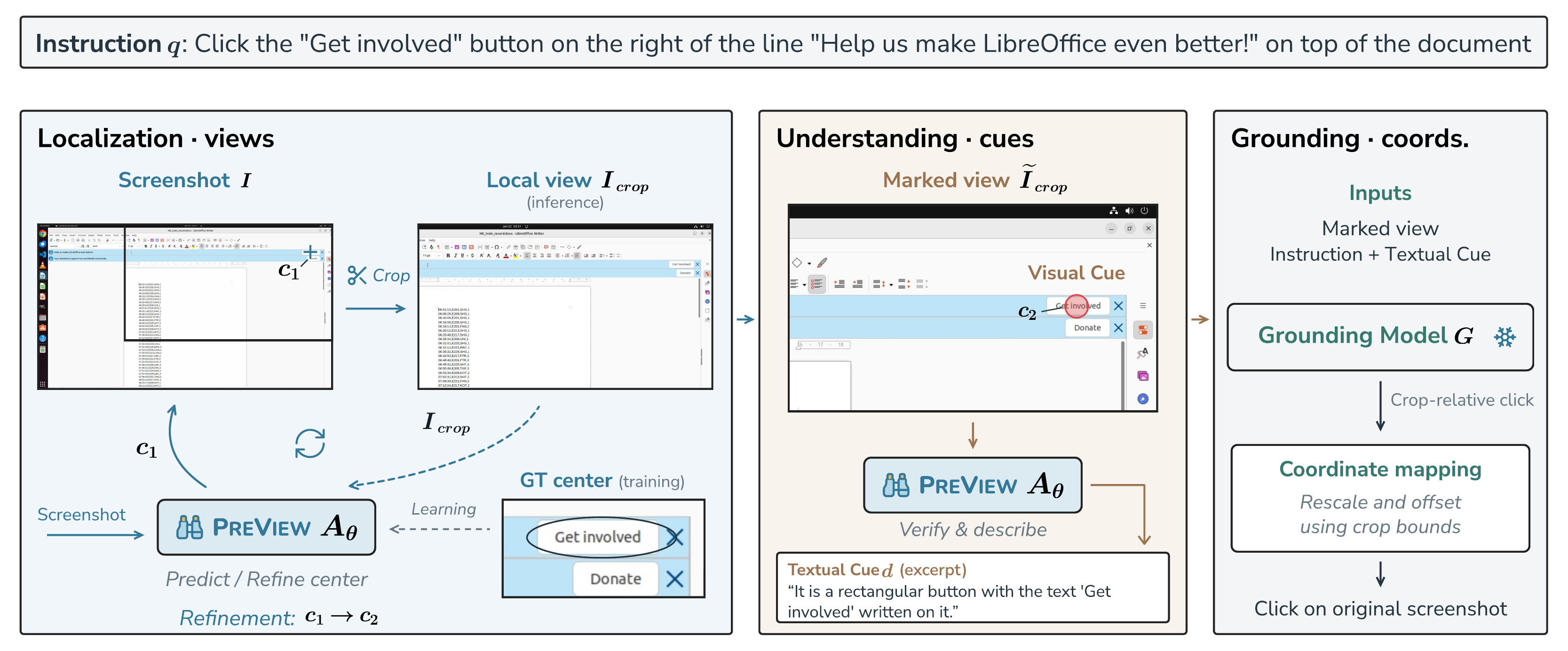}
    \caption{Overview of \method{}. For localization, the auxiliary model refines the initial center $c_1$ into $c_2$ to prepare a task-relevant view through cropping and enlargement. For understanding, it provides a visual marker with a location hint and generates a candidate description. The frozen grounding model uses the prepared view, instruction, and available cues in a single call. The predicted click is then mapped back to the original screenshot.}
    \label{fig:overview}
\end{figure*}
\section{Related Work}
\label{sec:related-work}

\paragraph{GUI grounding.}
GUI-specific supervision improves grounding in SeeClick, UGround, OS-Atlas, GroundCUA, and Phi-Ground \citep{cheng-etal-2024-seeclick,ICLR2025_4ca0e369,osatlas2024,groundcua2025,phiground2025}.
Architectural approaches include high-resolution encoding in CogAgent, visual-token selection in ShowUI, and coordinate-free action prediction in GUI-Actor \citep{cogagent2023,showui2024,guiactor2025}.
Aguvis and UI-TARS integrate grounding into broader visual agents \citep{aguvis2024,uitars2025}.
\method{} trains an auxiliary preprocessor while retaining the downstream model's parameters and click-prediction interface.

\paragraph{Zooming and test-time scaling.}
ReGround, ScreenSeekeR, RegionFocus, and ZoomClick improve observations through recropping or cascaded visual search \citep{screenspotpro2025,luo2025visualtesttimescalinggui,zoomclick2025}.
UI-Zoomer and AutoFocus use prediction uncertainty to guide zooming or region exploration \citep{uizoomer2026,autofocus2026}.
GTA1 uses a judge to select action proposals \citep{gta12025}.
\method{} assigns refinement and cue generation to an auxiliary model before one downstream grounding call.
This auxiliary inference adds computation.

\paragraph{Visual prompting and semantic augmentation.}
Spotlight, ScreenAI, and Ferret-UI connect UI regions with language and functional understanding \citep{spotlight2022,screenai2024,ferretui2024}.
Set-of-Mark supplies visual references \citep{yang2023setofmarkpromptingunleashesextraordinary}.
OmniParser detects and describes interactable elements \citep{omniparser2024}.
AdaZoom-GUI combines instruction refinement with zooming \citep{adazoomgui2026}.
\method{} prepares a task-relevant view and paired visual/textual cues.
Candidate verification and center consistency control which cues reach the frozen model.

\paragraph{Agent evaluation.}
Mind2Web studies cross-website generalization and auxiliary HTML filtering \citep{mind2web2023}.
VisualWebArena, OSWorld, and AndroidWorld evaluate interactive task completion \citep{visualwebarena2024,osworld2024,androidworld2024}.
Our five-benchmark suite isolates single-step grounding, separating input preparation from planning and multi-step recovery.
\section{Method}
\label{sec:method}

Given a screenshot $I$ and an instruction $q$, \method{} uses a trained auxiliary model $A_\theta$ to prepare inputs for a frozen grounding model $G$.
As shown in Figure~\ref{fig:overview}, it prepares a task-relevant view for localization and candidate-specific cues for target understanding.
The grounding model predicts the final click without parameter updates.

\subsection{Training}
\label{sec:training}

\paragraph{Center prediction.}
Given an unmarked screenshot and instruction, $A_\theta$ predicts the target-box center in image-relative coordinates.
This prediction determines where to crop rather than the final click.

\paragraph{Candidate verification and description.}
$A_\theta$ receives a local view, an instruction, and a marked candidate.
It learns from positive and negative examples to describe a matching candidate's identity or function and explicitly reject a mismatch.

\paragraph{Joint optimization.}
The two tasks share the parameters of $A_\theta$.
For each training example with multimodal input $x$ and target response $y$, we minimize the following loss:
\begin{equation*}
\mathcal{L}(\theta)=-\sum_{t=1}^{|y|}\log p_\theta(y_t\mid x,y_{<t}).
\end{equation*}
We compute the loss only over the target response tokens and update $A_\theta$ while keeping $G$ frozen.

\subsection{Inference}
\label{sec:inference}

\paragraph{Center-guided cropping.}
As shown in Figure~\ref{fig:overview}, the auxiliary model predicts an initial center $c_1$ from the full screenshot (\emph{Initial}), then refines it using a local view and the original instruction to obtain $c_2$ (\emph{Refine~1}).
We express both centers in normalized full-screenshot coordinates in $[0,1]^2$ using the mapping described below.
Inference stops at $c_2$, which determines the final unmarked view $I_{\mathrm{crop}}$ for cue generation and grounding.
A third prediction $c_3$ (\emph{Refine~2}) is used only for analysis.
For a screenshot of size $W\times H$, we construct a rectangle of size approximately $W\sqrt{\rho}\times H\sqrt{\rho}$ around the predicted center, where $\rho$ is the crop area ratio.
We shift the rectangle inside image boundaries when necessary and enlarge the resulting view.
Refinement uses a $50\%$-area crop enlarged by $\sqrt{2}$ per dimension (P50 $\rightarrow$ Full).
Description generation and grounding use a $12.5\%$-area crop enlarged $2\times$ (P12.5 $\rightarrow$ P50).
These budgets are relative to the original screenshot.

\paragraph{Visual and textual cues.}
The \emph{visual cue} pairs a translucent red circle at $c_2$ with a fixed location hint $h$ explaining the marker.
Although written in text, $h$ belongs to this cue because it instructs the model how to use the visual marker.
For the \emph{textual cue}, $A_\theta$ generates a description $d$ from the marked view and $q$.
The description identifies the candidate's identity, function, or relationships.
When cues are enabled and the candidate is accepted, the grounding input is
\begin{equation*}
\hat{s}=G(\widetilde I_{\mathrm{crop}},\,q\oplus h\oplus d),
\end{equation*}
where $\widetilde I_{\mathrm{crop}}$ is the marked view and $\oplus$ denotes instruction augmentation using the template in Appendix~\ref{sec:prompts}.
Here $\hat{s}$ is the grounding model's raw click prediction, distinct from the auxiliary model's center predictions.
The visual cue anchors the location, while the textual cue describes what is there.
A \texttt{REJECT} response indicates that $A_\theta$ judges the candidate inconsistent with $q$.
In this case, we withhold the textual cue $d$.

\paragraph{Center-consistency routing.}
We measure the displacement between the initial and refined centers in normalized image coordinates:
\begin{equation*}
\delta=\sqrt{(c_{1,x}-c_{2,x})^2+(c_{1,y}-c_{2,y})^2}.
\end{equation*}
When $\delta\leq\tau$ and the initial center is valid, the grounding model receives the marked view and available cues.
Otherwise, it receives the same unmarked view $I_{\mathrm{crop}}$ with the original instruction.
Even if the auxiliary model rejects the candidate, we retain the visual cue when consistency passes.

\paragraph{Coordinate mapping.}
We normalize the grounding model's raw click prediction $\hat{s}=(\hat{s}_x,\hat{s}_y)$ by its horizontal and vertical coordinate ranges $R_x$ and $R_y$, giving $\hat{u}=(\hat{s}_x/R_x,\,\hat{s}_y/R_y)\in[0,1]^2$ within the input view.
Let $(x_{\min},y_{\min})$ and $(x_{\max},y_{\max})$ denote the input view's top-left and bottom-right corners in the full screenshot after boundary adjustment.
We map the prediction back to a click $\hat{p}=(\hat{p}_x,\hat{p}_y)$ in the full screenshot as follows:
\begin{equation*}
\begin{aligned}
\hat{p}_x &= x_{\min}+(x_{\max}-x_{\min})\hat{u}_x, \\
\hat{p}_y &= y_{\min}+(y_{\max}-y_{\min})\hat{u}_y.
\end{aligned}
\end{equation*}

\section{Experiments}
\label{sec:experiments}
\begin{table*}[t]
    \centering
    \small
    \definecolor{gainbg}{HTML}{4FA77A}
    \newcommand{\gaincell}[2]{\begingroup\setlength{\fboxsep}{1pt}\colorbox{gainbg!#1}{\makebox[2.6em][c]{\textcolor{black}{$+#2$}}}\endgroup}
    \setlength{\tabcolsep}{5pt}
    \begin{tabular}{lccccccc}
    \toprule
    \textbf{Model} & \textbf{SSPro} & \textbf{OSW-G}$_{\mathrm{O}}$ & \textbf{OSW-G}$_{\mathrm{R}}$ & \textbf{MMB-GUI} & \textbf{SSv2} & \textbf{UI-V} & \textbf{Avg.} \\
    \midrule
    \rowcolor{black!5}
    \multicolumn{8}{l}{$\leq \mathbf{3}$\textbf{B}} \\
    UI-R1-E-3B & $33.5$ & -- & $48.8$ & $68.4$ & $88.6$ & $16.5$ & -- \\
    SE-GUI-3B & $35.9$ & -- & $46.1$ & $66.3$ & $86.8$ & $15.0$ & $50.0$ \\
    InfiGUI-R1-3B & $35.7$ & -- & $42.9$ & $70.6$ & $89.5$ & $17.8$ & $51.3$ \\
    GUI-G$^2$-3B & $36.4$ & -- & $53.5$ & $66.3$ & $87.6$ & $18.7$ & $52.5$ \\
    GUI-G1-3B & $37.1$ & -- & $49.5$ & $71.0$ & $89.5$ & $20.3$ & $53.5$ \\
    InfiGUI-G1-3B & $45.2$ & -- & $49.6$ & $73.4$ & $91.1$ & $22.0$ & $56.3$ \\
    Qwen2.5-VL-3B & $16.1$ & $27.3$ & -- & $60.8$ & $80.9$ & $6.3$ & $38.3$ \\
    JEDI-3B & $36.1$ & $50.9$ & $61.0$ & $66.5$ & $88.6$ & $18.7$ & $52.2$ \\
    GUI-Actor-3B & $42.2$ & -- & $48.9$ & $69.8$ & $91.0$ & $19.7$ & $54.3$ \\
    GroundNext-3B & $49.8$ & -- & $64.2$ & $77.1$ & $88.8$ & $62.1$ & $68.4$ \\
    \midrule[0.25pt]
    Qwen3-VL-2B & $43.1^{*}$ & $46.1$ & $62.1$ & $72.9$ & $91.4$ & $16.5$ & $57.2$ \\
    Qwen3-VL-2B (\textit{w.} \textsc{PreView}) & $\mathbf{59.3}^{*}$ & $\mathbf{55.0}$ & $\mathbf{66.8}$ & $77.0$ & $86.0$ & $33.7$ & $64.6$ \\
    $\Delta$ Improvement & \gaincell{53.7}{16.2} & \gaincell{40.8}{8.9} & \gaincell{33.5}{4.8} & \gaincell{32.3}{4.1} & $-5.4$ & \gaincell{55.3}{17.1} & \gaincell{38.1}{7.4} \\
    \midrule
    \rowcolor{black!5}
    \multicolumn{8}{l}{$>\mathbf{3}$\textbf{B}} \\
    SE-GUI-7B & $47.3$ & -- & $33.9$ & $34.5$ & $68.9$ & $16.7$ & $40.3$ \\
    UI-TARS-1.5-7B & $49.6$ & $52.8$ & $64.2$ & $64.3$ & $90.3$ & $20.8$ & $57.8$ \\
    GUI-G$^2$-7B & $47.5$ & -- & $61.9$ & $79.5$ & $93.3$ & $25.6$ & $61.7$ \\
    InfiGUI-G1-7B & $51.9$ & -- & $59.9$ & $80.8$ & $93.5$ & $26.1$ & $62.4$ \\
    GTA1-7B & $50.1$ & $55.1$ & $67.7$ & $79.4$ & $92.4$ & $25.7$ & $63.1$ \\
    PhiGround-4B-7C & $22.8$ & -- & $51.4$ & $60.3$ & $80.8$ & $20.5$ & $47.2$ \\
    Qwen2.5-VL-7B & $26.8$ & $31.4$ & -- & $33.9$ & $88.8$ & $0.9$ & $36.4$ \\
    OS-Atlas-7B & $18.9$ & $27.7$ & -- & $41.4$ & $85.1$ & $9.0$ & $36.4$ \\
    UGround-V1-7B & $16.5$ & $36.4$ & -- & $65.7$ & $87.6$ & $12.9$ & $43.8$ \\
    Aguvis-7B & $39.5$ & $38.7$ & -- & $45.7$ & $86.0$ & $13.7$ & $44.7$ \\
    GUI-Actor-7B & $44.6$ & -- & $47.0$ & $70.9$ & $92.1$ & $21.9$ & $55.3$ \\
    JEDI-7B & $39.5$ & $54.1$ & $63.8$ & $70.4$ & $91.7$ & $24.8$ & $56.1$ \\
    GroundNext-7B & $52.9$ & -- & $67.7$ & $81.1$ & $90.4$ & $60.3$ & $70.5$ \\
    \midrule[0.25pt]
    OS-Atlas-4B & $2.8^{*}$ & $10.5$ & $8.5$ & $26.6$ & $62.7$ & $7.3$ & $21.6$ \\
    OS-Atlas-4B (\textit{w.} \textsc{PreView}) & $28.2^{*}$ & $31.9$ & $25.9$ & $43.5$ & $61.2$ & $18.9$ & $35.5$ \\
    $\Delta$ Improvement & \gaincell{70.0}{25.4} & \gaincell{63.1}{21.5} & \gaincell{55.8}{17.4} & \gaincell{54.9}{16.9} & $-1.5$ & \gaincell{45.6}{11.6} & \gaincell{49.6}{13.9} \\
    \midrule[0.25pt]
    Qwen3-VL-4B & $59.3^{*}$ & $58.2$ & $70.6$ & $84.7$ & $94.0$ & $25.3$ & $66.8$ \\
    Qwen3-VL-4B (\textit{w.} \textsc{PreView}) & $\mathbf{63.7}^{*}$ & $\mathbf{59.9}$ & $70.4$ & $79.7$ & $87.1$ & $40.6$ & $68.3$ \\
    $\Delta$ Improvement & \gaincell{32.8}{4.4} & \gaincell{28.0}{1.7} & $-0.2$ & $-5.0$ & $-6.9$ & \gaincell{52.1}{15.3} & \gaincell{27.7}{1.5} \\
    \bottomrule
    \end{tabular}
    \caption{Grounding accuracy (\%) by model size.
    O and R denote original and refined OSW-G instructions, respectively.
    Bold marks \method{} scores that are column-best within each size group.
    $\Delta$ denotes percentage-point gains. $^{*}$ marks our SSPro evaluations. -- indicates unavailable or unverified scores.
    Reference averages follow GroundCUA. Our five-benchmark averages use OSW-G$_{\mathrm{R}}$.}
    \label{tab:baselines}
\end{table*}

\subsection{Datasets}
\paragraph{Training data.}
We construct training examples from 50,000 desktop-application screenshots in GroundCUA, using its element bounding boxes and official instruction-tuning annotations \citep{groundcua2025}.
Center supervision is the geometric target-box center, normalized to $[0,1000]^2$.
Positive descriptions come from the official instruction-tuning release, matched by screenshot and target element.
For each positive example, we randomly sample a candidate point within the target bounding box.
Negative examples use annotated non-target elements with the response \texttt{REJECT}.
The resulting mixture contains 50,000 center-prediction examples, 49,311 positive description examples, and 49,284 rejection examples.

\paragraph{Evaluation benchmarks.}
We adopt the five-benchmark suite used by \citet{groundcua2025}: ScreenSpot-Pro (SSPro; \citealp{screenspotpro2025}), OSWorld-G (OSW-G; \citealp{jedi2025}), MMBench-GUI L2 (MMB-GUI; \citealp{mmbenchgui2025}), ScreenSpot-v2 (SSv2; \citealp{osatlas2024}), and UI-Vision Element Grounding (UI-V; \citealp{uivision2025}).
OSW-G$_{\mathrm{O}}$ and OSW-G$_{\mathrm{R}}$ denote separate original/refined instruction conditions.
Table~\ref{tab:benchmark-statistics} in Appendix~\ref{sec:implementation} summarizes benchmark sizes.
Appendix~\ref{sec:benchmark-breakdown} reports category-level results for the two Qwen3-VL models.

\subsection{Evaluation Metrics}
Grounding accuracy measures whether the final click lies inside the annotated target box.
OSW-G also scores correct refusal when the requested target is absent.
For MMB-GUI, we follow the official evaluation code and report the mean of Basic and Advanced accuracies.
For the auxiliary model, \emph{direct hit} scores its predicted center, while \emph{center containment} tests whether the resulting $12.5\%$-area view contains the target-box center.
The refinement analysis evaluates Initial ($c_1$), Refine~1 ($c_2$), and Refine~2 ($c_3$) without description feedback or early stopping.
The main pipeline stops at $c_2$ (Section~\ref{sec:inference}).
Additional evaluation details are provided in Appendix~\ref{sec:implementation}.

\subsection{Baselines}
Table~\ref{tab:baselines} groups published results from \citet{groundcua2025} into $\leq 3$B and $>3$B models.
UI-R1-E's SSPro score follows its original paper \citep{uir12025}.
We evaluate \method{} with Qwen3-VL-2B-Instruct and Qwen3-VL-4B-Instruct \citep{qwen3vl2025}, and OS-Atlas-Base-4B \citep{osatlas2024}.
Each comparison group reports Full, \method{}, and their difference in percentage points.
Within these groups, Qwen3-VL Full OSW-G$_{\mathrm{O}}$ and ScreenSpot scores use official model cards.
Remaining Full scores are ours.

\subsection{Implementation Details}
We jointly train both auxiliary tasks from Qwen3.5-0.8B for two epochs in BF16 precision.
Language-model attention and feed-forward layers use LoRA \citep{lora2021}.
The multimodal merger is trainable and the vision encoder is frozen.
Table~\ref{tab:training-config} lists the training configuration.
We use vLLM for inference.

\begin{table}[t]
    \centering
    \small
    \begin{tabular}{ll}
        \toprule
        \textbf{Hyperparameter} & \textbf{Value} \\
        \midrule
        Epochs / precision & $2$ / BF16 \\
        Optimizer & AdamW \\
        Learning rate & $5\times10^{-5}$ \\
        Weight decay & $10^{-3}$ \\
        LR schedule / warmup & Linear / $3\%$ \\
        Micro / effective batch size & $4$ / $128$ \\
        Gradient accumulation & $32$ \\
        LoRA rank / $\alpha$ & $128$ / $256$ \\
        \bottomrule
    \end{tabular}
    \caption{Auxiliary-model training configuration.}
    \label{tab:training-config}
\end{table}

\paragraph{Inputs and routing.}
Center prediction receives the unmarked screenshot and instruction, and outputs coordinates in $[0,1000]^2$.
Candidate verification receives the instruction, a marked local view, and view-relative candidate coordinates on the same scale.
The view retains $12.5\%$ of the source area and is enlarged $2\times$.
The output is a description or \texttt{REJECT}.
Training descriptions are capped at 80 tokens, and auxiliary inference allows 128 output tokens.
At inference, all three grounding models use $\tau=0.02$ to select between a marked view with available cues and the same unmarked view.
Prompt templates appear in Appendix~\ref{sec:prompts}.

\paragraph{Visual budget ablation.}
We evaluate Qwen3-VL-2B-Instruct on OSW-G$_{\mathrm{R}}$, SSPro, and UI-V using unmarked views without descriptions.
We vary cropping area and upsampling while keeping each sample's refined center, instruction, and grounding settings fixed.
Pixel budgets denote output area relative to the full screenshot.
Equal-budget settings match processed visual-token counts per sample.
Configurations and evaluation coverage appear in Appendix~\ref{sec:implementation}.
\section{Results}
\label{sec:results}

\subsection{Grounding Performance}
Better inputs improve grounding across model families and can partly close the gap to a larger grounding model.
Table~\ref{tab:baselines} reports average gains of $7.4$, $13.9$, and $1.5$ percentage points for Qwen3-VL-2B, OS-Atlas-4B, and Qwen3-VL-4B, respectively.
The first two models improve on four benchmark families: SSPro, OSW-G, MMB-GUI, and UI-V.
With \method{}, Qwen3-VL-2B matches the 4B Full score on SSPro ($59.3\%$) and exceeds it on UI-V ($33.7\%$ versus $25.3\%$).

\method{} enables Qwen3-VL-2B and Qwen3-VL-4B to compete with larger 3B and 7B models, respectively.
This requires neither further training nor additional calls to the grounding models.
Both outperform all listed baselines of the corresponding larger size on SSPro and OSW-G Refined.
Their average accuracy is also competitive across benchmarks (Table~\ref{tab:baselines}).
These results suggest that improving localization and understanding through input preparation can partly bridge the gap in grounding-model size.

These aggregate gains vary across tasks and target types.
Qwen3-VL-4B gains $15.3$ points on UI-V but loses $5.0$ on MMB-GUI.
On SSPro, its icon accuracy rises from $32.5\%$ to $42.5\%$, versus $75.8\%$ to $76.8\%$ for text (Appendix~\ref{sec:benchmark-breakdown}).
The larger gain on icons suggests greater remaining benefit for these targets.
Within UI-V, both Qwen3-VL models improve on Basic, Functional, and Spatial instructions (Table~\ref{tab:breakdown-uivision}).
Spatial accuracy remains lowest, at $17.0\%$ for 2B and $26.0\%$ for 4B.
Interpreting spatial relationships remains a weakness for both models after preprocessing.

\begin{table}[t]
    \centering
    \footnotesize
    \setlength{\tabcolsep}{2pt}
    \definecolor{tfpositive}{HTML}{16845B}
    \begin{tabular}{@{}llccc@{}}
    \toprule
    \textbf{Method} & \textbf{Grounding model} & \textbf{Base} & \textbf{Enhanced} & $\boldsymbol{\Delta}$ \\
    \midrule
    ReGround & OS-Atlas-7B & $18.9$ & $40.2$ & \textcolor{tfpositive}{$+21.3$} \\
    ScreenSeekeR & OS-Atlas-7B & $18.9$ & $48.1$ & \textcolor{tfpositive}{$+29.2$} \\
    RegionFocus & Qwen2.5-VL-7B & $26.8$ & $32.1$ & \textcolor{tfpositive}{$+5.3$} \\
    ZoomClick & UI-Venus-7B & $50.3$ & $65.7$ & \textcolor{tfpositive}{$+15.4$} \\
    UI-Zoomer & UI-Venus-7B & $50.0$ & $61.8$ & \textcolor{tfpositive}{$+11.8$} \\
    AutoFocus & UI-Venus-7B & $50.3$ & $67.8$ & \textcolor{tfpositive}{$+17.5$} \\
    \midrule[0.25pt]
    \method{} & Qwen3-VL-2B & $43.1$ & $59.3$ & \textcolor{tfpositive}{$+16.2$} \\
    \method{} & OS-Atlas-4B & $2.8$ & $28.2$ & \textcolor{tfpositive}{$+25.4$} \\
    \method{} & Qwen3-VL-4B & $59.3$ & $63.7$ & \textcolor{tfpositive}{$+4.4$} \\
    \bottomrule
    \end{tabular}
    \caption{Comparison with training-free methods on SSPro. The Base and Enhanced columns show accuracy (\%). $\Delta$ is the gain in percentage points.}
    \label{tab:training-free-comparison}
\end{table}
\paragraph{Comparison with training-free methods.}
Table~\ref{tab:training-free-comparison} compares \method{} with training-free input and inference refinements on SSPro.
\method{} achieves gains of $16.2$, $25.4$, and $4.4$ points with \emph{one grounding-model call} for 2B, OS-Atlas-4B, and 4B, respectively.
The lightweight auxiliary model handles localization refinement and cue generation, so the grounding model need not repeat its search.
Integration requires neither logits nor hidden states.
It preserves the grounder's input interface and decoding procedure, adding only preprocessing and coordinate mapping.

\begin{figure}[t]
    \centering
    \includegraphics[width=\columnwidth]{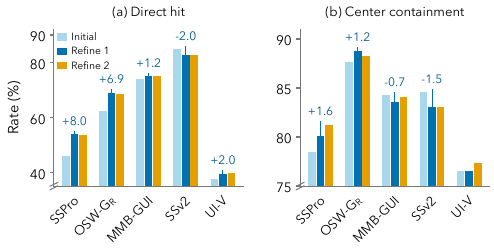}
    \caption{Auxiliary-model direct hit (a) and center containment (b). Initial, Refine~1, and Refine~2 correspond to $c_1$, $c_2$, and $c_3$, respectively.}
    \label{fig:center-rounds}
\end{figure}
\subsection{Auxiliary Model Performance}
\label{sec:auxiliary-performance}

A center prediction can be useful for selecting a view even when it is not accurate enough for clicking.
On SSPro, Refine~1 retains the target center in $80.14\%$ of views but directly hits the target in only $53.83\%$ of cases (Figure~\ref{fig:center-rounds}).
One refinement raises mean direct hit from $60.83\%$ to $64.02\%$ at $c_2$.
The analysis-only $c_3$ yields $63.93\%$, providing no further gain.
We therefore stop at $c_2$ rather than adding another auxiliary call.

\begin{table}[t]
\centering
\small
\begin{tabular}{lrrr}
\toprule
 & \multicolumn{2}{c}{Direct hit (\%)} & \\
\cmidrule(lr){2-3}
Benchmark & Accept & Reject & $N_{\mathrm{Reject}}$ \\
\midrule
SSPro & $62.0$ & $18.6$ & $295$ \\
OSW-G$_{\mathrm{R}}$ & $77.2$ & $25.4$ & $71$ \\
MMB-GUI & $82.0$ & $41.3$ & $625$ \\
SSv2 & $90.0$ & $63.4$ & $344$ \\
UI-V & $49.6$ & $12.8$ & $1520$ \\
\midrule
Weighted Avg. & $66.5$ & $26.1$ & $2855$ \\
\bottomrule
\end{tabular}
\caption{Auxiliary-model direct hit by verification response, on target-present examples. Weighted Avg. weights each benchmark by its accepted or rejected sample count for the corresponding column.}
\label{tab:rejection}
\end{table}
\paragraph{Candidate rejection.}
A rejected candidate can still be correctly located.
Rejected candidates have lower sample-weighted direct hit than accepted ones ($26.1\%$ versus $66.5\%$).
On SSv2, rejected candidates reach $63.4\%$ direct hit (Table~\ref{tab:rejection}).
On $1{,}163$ rejected candidates passing consistency, keeping visual cues produces $28$ and $12$ more correct clicks for 2B and 4B than View-only.
These gains show that spatial guidance can remain useful even when the auxiliary model rejects the candidate.

\begin{table*}[t]
    \centering
    \small
    \setlength{\tabcolsep}{3.4pt}
    \renewcommand{\arraystretch}{0.95}
    \begin{tabular*}{\textwidth}{@{\extracolsep{\fill}}llccccccccc@{}}
    \toprule
    \textbf{Model} & \textbf{Input} & \textbf{Visual Cue} & \textbf{Textual Cue} & \textbf{SSPro} & \textbf{OSW-G}$_{\mathrm{O}}$ & \textbf{OSW-G}$_{\mathrm{R}}$ & \textbf{MMB-GUI} & \textbf{SSv2} & \textbf{UI-V} & \textbf{Avg.} \\
    \midrule
    \multirow{4}{*}{Qwen3-VL-2B}
      & Full & -- & -- & $43.1$ & $46.1$ & $62.1$ & $72.9$ & $91.4$ & $16.5$ & $57.2$ \\
      & View & -- & -- & $56.9$ & $52.7$ & $66.3$ & $75.7$ & $84.8$ & $26.0$ & $61.9$ \\
      & View & $\checkmark$ & -- & $55.6$ & $54.8$ & $65.8$ & $76.4$ & $85.2$ & $32.7$ & $63.1$ \\
      & View & $\checkmark$ & $\checkmark$ & $59.3$ & $55.0$ & $66.8$ & $77.0$ & $86.0$ & $33.7$ & $64.6$ \\
    \midrule[0.25pt]
    \multirow{4}{*}{OS-Atlas-4B}
      & Full & -- & -- & $2.8$ & $10.5$ & $8.5$ & $26.6$ & $62.7$ & $7.3$ & $21.6$ \\
      & View & -- & -- & $21.7$ & $27.1$ & $24.1$ & $37.8$ & $63.0$ & $13.0$ & $31.9$ \\
      & View & $\checkmark$ & -- & $25.2$ & $29.4$ & $22.2$ & $41.5$ & $63.0$ & $16.4$ & $33.7$ \\
      & View & $\checkmark$ & $\checkmark$ & $28.2$ & $31.9$ & $25.9$ & $43.5$ & $61.2$ & $18.9$ & $35.5$ \\
    \midrule[0.25pt]
    \multirow{4}{*}{Qwen3-VL-4B}
      & Full & -- & -- & $59.3$ & $58.2$ & $70.6$ & $84.7$ & $94.0$ & $25.3$ & $66.8$ \\
      & View & -- & -- & $65.0$ & $59.4$ & $71.5$ & $79.9$ & $86.2$ & $35.9$ & $67.7$ \\
      & View & $\checkmark$ & -- & $64.0$ & $60.1$ & $70.2$ & $80.0$ & $86.7$ & $39.5$ & $68.1$ \\
      & View & $\checkmark$ & $\checkmark$ & $63.7$ & $59.9$ & $70.4$ & $79.7$ & $87.1$ & $40.6$ & $68.3$ \\
    \bottomrule
    \end{tabular*}
    \caption{View and cue ablation (accuracy, \%). Full uses the original screenshot. Avg.\ uses OSW-G$_{\mathrm{R}}$.}
    \label{tab:cue-ablation}
\end{table*}
\subsection{View and Cue Ablation}
\label{sec:cue-ablation}
\paragraph{Task-relevant views.}
View preparation supplies much of the improvement before any cue is added (Table~\ref{tab:cue-ablation}).
For 2B on SSPro, View raises accuracy from $43.1\%$ to $56.9\%$, already close to $59.3\%$ with complete cues.
Cues contribute more on UI-V, adding $7.7$ points beyond View compared with $2.5$ on SSPro.
View and cue gains depend on whether a task primarily requires region selection or target interpretation.

\paragraph{Visual and textual cues.}
Indicating a location alone does not always resolve the target's identity.
On SSPro, 2B falls from $56.9\%$ with View to $55.6\%$ with the marker and hint, then reaches $59.3\%$ when the description is added.
Here, the improvement comes from adding the description to the marker and hint.
Combined cues improve average accuracy beyond View for all three models.
For 4B on SSPro, accuracy falls from $65.0\%$ with View to $64.0\%$ with visual cues and $63.7\%$ with both cue types.
The loss begins with the marker and hint, and adding a description does not recover it.
One possible explanation is that inaccurate guidance can redirect a prediction that the model would make correctly from the view alone.
On UI-V, the same 4B model benefits from combined cues, improving from $35.9\%$ to $40.6\%$.
Cue usefulness therefore depends on the task as well as the grounding model.

\begin{figure}[t]
    \centering
    \includegraphics[width=\columnwidth]{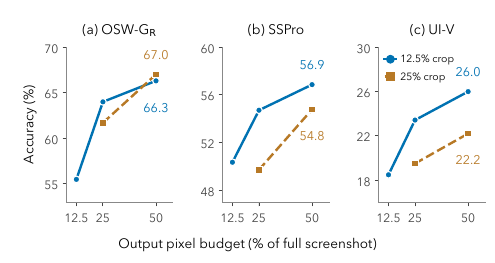}
    \caption{Effects of cropping area and enlargement on Qwen3-VL-2B at different visual budgets.}
    \label{fig:visual-budget}
\end{figure}
\subsection{Visual Budget Ablation}
\label{sec:visual-budget}
Selecting useful content and enlarging it make complementary contributions (Figure~\ref{fig:visual-budget}).
At the same $25\%$ pixel budget, enlarging a $12.5\%$-area crop beats using a native $25\%$-area crop on all three benchmarks with identical processed visual-token counts.
The gain therefore depends on how the budget is allocated, not simply on supplying more tokens.
For a fixed $12.5\%$-area crop, doubling both dimensions adds $6.5$--$10.8$ points, showing the additional value of enlargement.
At the same $50\%$ visual budget, smaller crops perform better on SSPro and UI-V, while the wider crop improves OSW-G$_{\mathrm{R}}$ by $0.71$ points.

\begin{figure}[t]
    \centering
    \includegraphics[width=0.75\columnwidth]{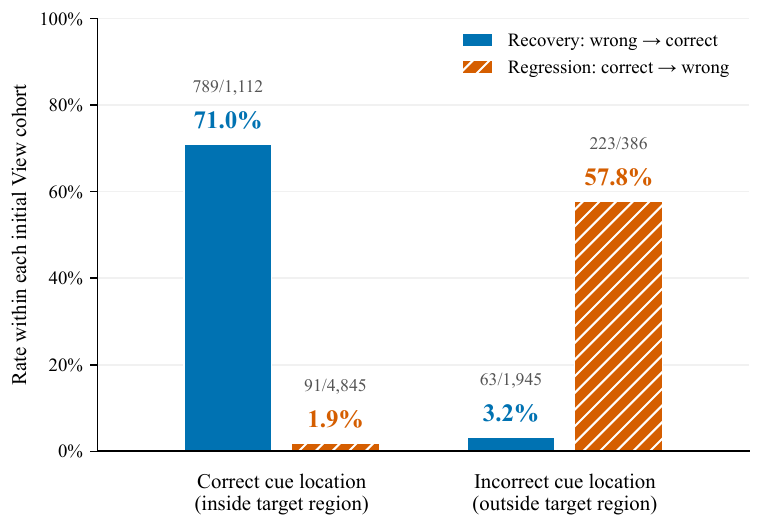}
    \caption{Cue recovery and regression under correct and incorrect cue locations for Qwen3-VL-2B.}
    \label{fig:cue-recovery}
\end{figure}
\subsection{Cue Reliability and Failure Analysis}
\label{sec:cue-reliability}
Cues help most when their location is correct, but can override a correct grounding decision when misplaced.
Figure~\ref{fig:cue-recovery} compares View-only predictions with those using paired visual and textual cues on 8,288 target-present 2B examples passing the consistency filter.
Recovery measures how often cues correct initially wrong View predictions, whereas regression measures how often they turn initially correct predictions into errors.
For this analysis, we group examples by whether the cue lies inside the annotated target.

When the cue lies inside the annotated target, it corrects $71.0\%$ of View errors and disrupts only $1.9\%$ of correct predictions.
Outside the target, these rates become $3.2\%$ and $57.8\%$.
Although the filter suppresses cues with inconsistent center predictions, passing it does not guarantee a correct cue location.

\paragraph{Where regressions arise.}
On MMB-GUI, 4B falls from $84.7\%$ with Full to $79.9\%$ with View and $79.7\%$ with cues.
Most of this loss precedes cue injection.
On SSv2, 2B similarly falls from $91.4\%$ to $84.8\%$ with View and recovers only to $86.0\%$ with cues.
Consistency-based fallback removes cues but keeps the same view.
Cue filtering therefore cannot repair errors caused by excluding the target or necessary context.

The 2B model improves across Windows, macOS, and Linux but regresses on both iOS subsets (Table~\ref{tab:breakdown-mmbench}).
This pattern is consistent with our desktop-only auxiliary training.
Platform alone cannot explain why Windows/Advanced improves for 2B but declines for 4B.
Higher Full accuracy leaves fewer errors to fix and more correct predictions to disrupt.
On OSW-G$_{\mathrm{R}}$, 4B improves from $392$ to $395$ correct target-present predictions.
Correct refusals on target-absent tasks fall from $6$ to $2$, offsetting this gain and slightly lowering the overall score (Table~\ref{tab:breakdown-osworld}).

\paragraph{Qualitative case studies.}
The examples in Figure~\ref{fig:case-studies} (Appendix~\ref{sec:case-studies}) distinguish seeing a control from identifying its role.
View alone recovers the Cut button in (a), but still confuses Edit with its dropdown in (b).
Cues resolve the latter by specifying the dropdown's function.
The comparison illustrates why clearer local evidence can suffice for one instruction while another needs help interpreting adjacent controls.

In (c), the center lies inside the date-and-time control, but an accepted description refers to a neighboring notification icon.
Adding cues moves a correct click to that icon.
In (d), cropping excludes the target cell U42 entirely.
These cases reveal two ways that input preparation can harm grounding.
Cropping can remove necessary visual evidence, while misleading cues can override an otherwise correct prediction.

\begin{table}[t]
    \centering
    \small
    \setlength{\tabcolsep}{8pt}
    \definecolor{gngain}{HTML}{16845B}
    \definecolor{gnloss}{HTML}{C0392B}
    \newcommand{\gnup}{\textcolor{gngain}{\scriptsize$\uparrow$}}
    \newcommand{\gndown}{\textcolor{gnloss}{\scriptsize$\downarrow$}}
    \begin{tabular}{lcc}
    \toprule
    \textbf{Benchmark} & \textbf{Full} & \textbf{View} \\
    \midrule
    SSPro & \textcolor{black!55}{$52.9$} & $59.0$\,\gnup \\
    OSW-G$_{\mathrm{O}}$ & \textcolor{black!55}{$56.2$} & $57.6$\,\gnup \\
    OSW-G$_{\mathrm{R}}$ & \textcolor{black!55}{$67.7$} & $68.3$\,\gnup \\
    MMB-GUI & \textcolor{black!55}{$81.1$} & $76.9$\,\gndown \\
    SSv2 & \textcolor{black!55}{$90.4$} & $81.8$\,\gndown \\
    UI-V & \textcolor{black!55}{$60.3$} & $54.5$\,\gndown \\
    \bottomrule
    \end{tabular}
    \caption{GroundNext-7B with View (accuracy, \%). Arrows compare with Full.}
    \label{tab:groundnext-preview}
\end{table}
\subsection{Transfer with Overlapping Training Data}
Input preparation can help even when the grounding model was trained on the data used to train the auxiliary model.
\method{} constructs 148,595 training examples from a subset of GroundCUA, only $21\%$ as many as the 700,000 examples used for GroundNext's supervised fine-tuning \citep{groundcua2025}.
With View, GroundNext's accuracy rises from $52.9\%$ to $59.0\%$ on SSPro and from $67.7\%$ to $68.3\%$ on OSW-G$_{\mathrm{R}}$ (Table~\ref{tab:groundnext-preview}).
These gains suggest that GroundNext can still benefit from task-relevant views after being trained on GroundCUA, the same source used to train \method{}.

\section{Conclusion}
We introduced \method{}, an auxiliary preprocessor that improves GUI grounding without updating the downstream model.
Center-guided views support localization, while visual and textual cues support target understanding.
Across five benchmarks and three grounding models, it improves reported average accuracy by $1.5$--$13.9$ points.
These gains require only one call to the frozen grounding model.
Ablations demonstrate complementary benefits from views and cues, while error analysis identifies target exclusion and misleading descriptions as failure modes.
Future work should adapt view selection and cue use based on estimated target coverage and cue reliability.
\FloatBarrier
\section*{Limitations}

Cropping may discard targets or necessary context, while incorrect cues may mislead the grounding model.
Training on desktop applications limits generalization to mobile interfaces, and offline evaluation does not establish reliability in multi-step interaction.
Auxiliary inference adds computational cost and latency.
We report a single run per configuration, so results may be influenced by initialization noise.
\section*{Ethical Considerations}

The datasets used in this study are publicly available and were obtained through their official release channels.
Incorrect grounding predictions can lead to unintended GUI operations, potentially modifying or deleting information or triggering unwanted actions.
Outputs should therefore be carefully reviewed before use, with human confirmation for consequential operations.

\bibliography{preview}

\begin{thebibliography}{32}
\providecommand{\natexlab}[1]{#1}

\bibitem[{Baechler et~al.(2024)Baechler, Sunkara, Wang, Zubach, Mansoor, Etter, C{\u a}rbune, Lin, Chen, and Sharma}]{screenai2024}
Gilles Baechler, Srinivas Sunkara, Maria Wang, Fedir Zubach, Hassan Mansoor, Vincent Etter, Victor C{\u a}rbune, Jason Lin, Jindong Chen, and Abhanshu Sharma. 2024.
\newblock \href {https://arxiv.org/abs/2402.04615} {{ScreenAI: A Vision-Language Model for UI and Infographics Understanding}}.
\newblock \emph{Preprint}, arXiv:2402.04615.

\bibitem[{Bai et~al.(2025)Bai, Cai, Chen, Chen, Chen, Cheng, Deng, Ding, Gao, Ge, Ge, Guo, Huang, Huang, Huang, Hui, Jiang, Li, Li, Li, Li, Lin, Lin, Liu, Liu, Liu, Liu, Liu, Liu, Lu, Luo, Lv, Men, Meng, Ren, Ren, Song, Sun, Tang, Tu, Wan, Wang, Wang, Wang, Wang, Xie, Xu, Xu, Xu, Yang, Yang, Yang, Yang, Yu, Zhang, Zhang, Zhang, Zheng, Zhong, Zhou, Zhou, Zhou, Zhu, and Zhu}]{qwen3vl2025}
Shuai Bai, Yuxuan Cai, Ruizhe Chen, Keqin Chen, Xionghui Chen, Zesen Cheng, Lianghao Deng, Wei Ding, Chang Gao, Chunjiang Ge, Wenbin Ge, Zhifang Guo, Qidong Huang, Jie Huang, Fei Huang, Binyuan Hui, Shutong Jiang, Zhaohai Li, Mingsheng Li, and 45 others. 2025.
\newblock \href {https://arxiv.org/abs/2511.21631} {{Qwen3-VL Technical Report}}.
\newblock \emph{Preprint}, arXiv:2511.21631.

\bibitem[{Cheng et~al.(2024)Cheng, Sun, Chu, Xu, Li, Zhang, and Wu}]{cheng-etal-2024-seeclick}
Kanzhi Cheng, Qiushi Sun, Yougang Chu, Fangzhi Xu, Yantao Li, Jianbing Zhang, and Zhiyong Wu. 2024.
\newblock \href {https://doi.org/10.18653/v1/2024.acl-long.505} {{S}ee{C}lick: Harnessing {GUI} grounding for advanced visual {GUI} agents}.
\newblock In \emph{Proceedings of the 62nd Annual Meeting of the Association for Computational Linguistics (Volume 1: Long Papers)}, pages 9313--9332, Bangkok, Thailand. Association for Computational Linguistics.

\bibitem[{Deng et~al.(2023)Deng, Gu, Zheng, Chen, Stevens, Wang, Sun, and Su}]{mind2web2023}
Xiang Deng, Yu~Gu, Boyuan Zheng, Shijie Chen, Samuel Stevens, Boshi Wang, Huan Sun, and Yu~Su. 2023.
\newblock \href {https://arxiv.org/abs/2306.06070} {{Mind2Web: Towards a Generalist Agent for the Web}}.
\newblock \emph{Preprint}, arXiv:2306.06070.

\bibitem[{Feizi et~al.(2025)Feizi, Nayak, Jian, Lin, Li, Awal, Lù, Obando-Ceron, Rodriguez, Chapados, Vazquez, Romero-Soriano, Rabbany, Taslakian, Pal, Gella, and Rajeswar}]{groundcua2025}
Aarash Feizi, Shravan Nayak, Xiangru Jian, Kevin~Qinghong Lin, Kaixin Li, Rabiul Awal, Xing~Han Lù, Johan Obando-Ceron, Juan~A. Rodriguez, Nicolas Chapados, David Vazquez, Adriana Romero-Soriano, Reihaneh Rabbany, Perouz Taslakian, Christopher Pal, Spandana Gella, and Sai Rajeswar. 2025.
\newblock \href {https://doi.org/10.48550/ARXIV.2511.07332} {{Grounding Computer Use Agents on Human Demonstrations}}.
\newblock \emph{arXiv preprint}.

\bibitem[{Gou et~al.(2025)Gou, Wang, Zheng, Xie, Chang, Shu, Sun, and Su}]{ICLR2025_4ca0e369}
Boyu Gou, Ruohan Wang, Boyuan Zheng, Yanan Xie, Cheng Chang, Yiheng Shu, Huan Sun, and Yu~Su. 2025.
\newblock \href {https://proceedings.iclr.cc/paper_files/paper/2025/file/4ca0e369689dadb25a5345ba9755ad6f-Paper-Conference.pdf} {Navigating the digital world as humans do: Universal visual grounding for {GUI} agents}.
\newblock In \emph{International Conference on Learning Representations}, volume 2025, pages 30851--30883.

\bibitem[{Hong et~al.(2024)Hong, Wang, Lv, Xu, Yu, Ji, Wang, Wang, Zhang, Li, Xu, Dong, Ding, and Tang}]{cogagent2023}
Wenyi Hong, Weihan Wang, Qingsong Lv, Jiazheng Xu, Wenmeng Yu, Junhui Ji, Yan Wang, Zihan Wang, Yuxuan Zhang, Juanzi Li, Bin Xu, Yuxiao Dong, Ming Ding, and Jie Tang. 2024.
\newblock \href {https://arxiv.org/abs/2312.08914} {{CogAgent: A Visual Language Model for GUI Agents}}.
\newblock \emph{Preprint}, arXiv:2312.08914.

\bibitem[{Hu et~al.(2021)Hu, Shen, Wallis, Allen-Zhu, Li, Wang, Wang, and Chen}]{lora2021}
Edward~J. Hu, Yelong Shen, Phillip Wallis, Zeyuan Allen-Zhu, Yuanzhi Li, Shean Wang, Lu~Wang, and Weizhu Chen. 2021.
\newblock \href {https://arxiv.org/abs/2106.09685} {{LoRA: Low-Rank Adaptation of Large Language Models}}.
\newblock \emph{Preprint}, arXiv:2106.09685.

\bibitem[{Jiang et~al.(2025)Jiang, Xie, Li, Zu, Li, Qiu, Pei, Ma, Huang, Wang, and Liu}]{zoomclick2025}
Zhiyuan Jiang, Shenghao Xie, Wenyi Li, Wenqiang Zu, Peihang Li, Jiahao Qiu, Siqi Pei, Lei Ma, Tiejun Huang, Mengdi Wang, and Shilong Liu. 2025.
\newblock \href {https://doi.org/10.48550/ARXIV.2512.05941} {{Zoom in, Click out: Unlocking and Evaluating the Potential of Zooming for GUI Grounding}}.
\newblock \emph{arXiv preprint}.

\bibitem[{Koh et~al.(2024)Koh, Lo, Jang, Duvvur, Lim, Huang, Neubig, Zhou, Salakhutdinov, and Fried}]{visualwebarena2024}
Jing~Yu Koh, Robert Lo, Lawrence Jang, Vikram Duvvur, Ming~Chong Lim, Po-Yu Huang, Graham Neubig, Shuyan Zhou, Ruslan Salakhutdinov, and Daniel Fried. 2024.
\newblock \href {https://arxiv.org/abs/2401.13649} {{VisualWebArena: Evaluating Multimodal Agents on Realistic Visual Web Tasks}}.
\newblock \emph{Preprint}, arXiv:2401.13649.

\bibitem[{Li and Li(2023)}]{spotlight2022}
Gang Li and Yang Li. 2023.
\newblock \href {https://arxiv.org/abs/2209.14927} {{Spotlight: Mobile UI Understanding using Vision-Language Models with a Focus}}.
\newblock \emph{Preprint}, arXiv:2209.14927.

\bibitem[{Li et~al.(2025)Li, Meng, Lin, Luo, Tian, Ma, Huang, and Chua}]{screenspotpro2025}
Kaixin Li, Ziyang Meng, Hongzhan Lin, Ziyang Luo, Yuchen Tian, Jing Ma, Zhiyong Huang, and Tat-Seng Chua. 2025.
\newblock \href {https://doi.org/10.48550/ARXIV.2504.07981} {{ScreenSpot-Pro}: {GUI} grounding for professional high-resolution computer use}.
\newblock \emph{arXiv preprint}.

\bibitem[{Lin et~al.(2024)Lin, Li, Gao, Yang, Wu, Bai, Lei, Wang, and Shou}]{showui2024}
Kevin~Qinghong Lin, Linjie Li, Difei Gao, Zhengyuan Yang, Shiwei Wu, Zechen Bai, Stan~Weixian Lei, Lijuan Wang, and Mike~Zheng Shou. 2024.
\newblock \href {https://arxiv.org/abs/2411.17465} {{ShowUI: One Vision-Language-Action Model for GUI Visual Agent}}.
\newblock \emph{Preprint}, arXiv:2411.17465.

\bibitem[{Lu et~al.(2024)Lu, Yang, Shen, and Awadallah}]{omniparser2024}
Yadong Lu, Jianwei Yang, Yelong Shen, and Ahmed Awadallah. 2024.
\newblock \href {https://doi.org/10.48550/ARXIV.2408.00203} {{OmniParser for Pure Vision Based GUI Agent}}.
\newblock \emph{arXiv preprint}.

\bibitem[{Lu et~al.(2025)Lu, Chai, Guo, Yin, Liu, Wang, Xiao, Ren, Xiong, and Li}]{uir12025}
Zhengxi Lu, Yuxiang Chai, Yaxuan Guo, Xi~Yin, Liang Liu, Hao Wang, Han Xiao, Shuai Ren, Guanjing Xiong, and Hongsheng Li. 2025.
\newblock \href {https://doi.org/10.48550/arXiv.2503.21620} {{UI-R1: Enhancing Efficient Action Prediction of GUI Agents by Reinforcement Learning}}.
\newblock \emph{arXiv preprint}.

\bibitem[{Luo et~al.(2025)Luo, Logeswaran, Johnson, and Lee}]{luo2025visualtesttimescalinggui}
Tiange Luo, Lajanugen Logeswaran, Justin Johnson, and Honglak Lee. 2025.
\newblock \href {https://arxiv.org/abs/2505.00684} {Visual test-time scaling for {GUI} agent grounding}.
\newblock \emph{Preprint}, arXiv:2505.00684.

\bibitem[{Nayak et~al.(2025)Nayak, Jian, Lin, Rodriguez, Kalsi, Awal, Chapados, Özsu, Agrawal, Vazquez, Pal, Taslakian, Gella, and Rajeswar}]{uivision2025}
Shravan Nayak, Xiangru Jian, Kevin~Qinghong Lin, Juan~A. Rodriguez, Montek Kalsi, Rabiul Awal, Nicolas Chapados, M.~Tamer Özsu, Aishwarya Agrawal, David Vazquez, Christopher Pal, Perouz Taslakian, Spandana Gella, and Sai Rajeswar. 2025.
\newblock \href {https://doi.org/10.48550/ARXIV.2503.15661} {{UI-Vision: A Desktop-centric GUI Benchmark for Visual Perception and Interaction}}.
\newblock \emph{arXiv preprint}.

\bibitem[{Pei et~al.(2026)Pei, Tang, Duan, Chen, Li, Huang, Jing, Yan, Zhang, Jiang, Zhang, and Lu}]{adazoomgui2026}
Siqi Pei, Liang Tang, Tiaonan Duan, Long Chen, Shuxian Li, Kaer Huang, Yanzhe Jing, Yiqiang Yan, Bo~Zhang, Chenghao Jiang, Borui Zhang, and Jiwen Lu. 2026.
\newblock \href {https://doi.org/10.48550/ARXIV.2603.17441} {{AdaZoom-GUI: Adaptive Zoom-based GUI Grounding with Instruction Refinement}}.
\newblock \emph{arXiv preprint}.

\bibitem[{Qin et~al.(2025)Qin, Ye, Fang, Wang, Liang, Tian, Zhang, Li, Li, Huang, Zhong, Li, Yang, Miao, Lin, Liu, Jiang, Ma, Li, Xiao, Cai, Li, Zheng, Jin, Li, Zhou, Wang, Chen, Li, Yang, Liu, Lin, Peng, Liu, and Shi}]{uitars2025}
Yujia Qin, Yining Ye, Junjie Fang, Haoming Wang, Shihao Liang, Shizuo Tian, Junda Zhang, Jiahao Li, Yunxin Li, Shijue Huang, Wanjun Zhong, Kuanye Li, Jiale Yang, Yu~Miao, Woyu Lin, Longxiang Liu, Xu~Jiang, Qianli Ma, Jingyu Li, and 16 others. 2025.
\newblock \href {https://doi.org/10.48550/ARXIV.2501.12326} {{UI-TARS: Pioneering Automated GUI Interaction with Native Agents}}.
\newblock \emph{arXiv preprint}.

\bibitem[{Rawles et~al.(2025)Rawles, Clinckemaillie, Chang, Waltz, Lau, Fair, Li, Bishop, Li, Campbell-Ajala, Toyama, Berry, Tyamagundlu, Lillicrap, and Riva}]{androidworld2024}
Christopher Rawles, Sarah Clinckemaillie, Yifan Chang, Jonathan Waltz, Gabrielle Lau, Marybeth Fair, Alice Li, William Bishop, Wei Li, Folawiyo Campbell-Ajala, Daniel Toyama, Robert Berry, Divya Tyamagundlu, Timothy Lillicrap, and Oriana Riva. 2025.
\newblock \href {https://arxiv.org/abs/2405.14573} {{AndroidWorld: A Dynamic Benchmarking Environment for Autonomous Agents}}.
\newblock \emph{Preprint}, arXiv:2405.14573.

\bibitem[{Tang et~al.(2026)Tang, Chen, Lu, Chen, Nong, Jiang, Xu, Lu, Xiao, Zhuang, and Shen}]{uizoomer2026}
Fei Tang, Bofan Chen, Zhengxi Lu, Tongbo Chen, Songqin Nong, Tao Jiang, Wenhao Xu, Weiming Lu, Jun Xiao, Yueting Zhuang, and Yongliang Shen. 2026.
\newblock \href {https://doi.org/10.48550/ARXIV.2604.14113} {{UI-Zoomer: Uncertainty-Driven Adaptive Zoom-In for GUI Grounding}}.
\newblock \emph{arXiv preprint}.

\bibitem[{Wang et~al.(2025)Wang, Wu, Xie, Ding, Yang, Li, Liu, Li, Dong, Chen, Wang, Zhao, Chen, Duan, Xie, Yang, Su, Yu, Huang, Liu, Zhang, Zhang, Yue, Su, Zhu, Shen, Dai, and Wang}]{mmbenchgui2025}
Xuehui Wang, Zhenyu Wu, JingJing Xie, Zichen Ding, Bowen Yang, Zehao Li, Zhaoyang Liu, Qingyun Li, Xuan Dong, Zhe Chen, Weiyun Wang, Xiangyu Zhao, Jixuan Chen, Haodong Duan, Tianbao Xie, Chenyu Yang, Shiqian Su, Yue Yu, Yuan Huang, and 9 others. 2025.
\newblock \href {https://doi.org/10.48550/ARXIV.2507.19478} {{MMBench-GUI: Hierarchical Multi-Platform Evaluation Framework for GUI Agents}}.
\newblock \emph{arXiv preprint}.

\bibitem[{Wu et~al.(2025)Wu, Cheng, Yang, Zhang, Yang, Jiang, Mu, Peng, Qiao, Tan, Qin, Liden, Lin, Zhang, Zhang, Zhang, Zhang, and Gao}]{guiactor2025}
Qianhui Wu, Kanzhi Cheng, Rui Yang, Chaoyun Zhang, Jianwei Yang, Huiqiang Jiang, Jian Mu, Baolin Peng, Bo~Qiao, Reuben Tan, Si~Qin, Lars Liden, Qingwei Lin, Huan Zhang, Tong Zhang, Jianbing Zhang, Dongmei Zhang, and Jianfeng Gao. 2025.
\newblock \href {https://doi.org/10.48550/ARXIV.2506.03143} {{GUI-Actor: Coordinate-Free Visual Grounding for GUI Agents}}.
\newblock \emph{arXiv preprint}.

\bibitem[{Wu et~al.(2024)Wu, Wu, Xu, Wang, Sun, Jia, Cheng, Ding, Chen, Liang, and Qiao}]{osatlas2024}
Zhiyong Wu, Zhenyu Wu, Fangzhi Xu, Yian Wang, Qiushi Sun, Chengyou Jia, Kanzhi Cheng, Zichen Ding, Liheng Chen, Paul~Pu Liang, and Yu~Qiao. 2024.
\newblock \href {https://doi.org/10.48550/ARXIV.2410.23218} {{OS-ATLAS: A Foundation Action Model for Generalist GUI Agents}}.
\newblock \emph{arXiv preprint}.

\bibitem[{Xie et~al.(2025)Xie, Deng, Li, Yang, Wu, Chen, Hu, Wang, Xu, Wang, Xu, Wang, Sahoo, Yu, and Xiong}]{jedi2025}
Tianbao Xie, Jiaqi Deng, Xiaochuan Li, Junlin Yang, Haoyuan Wu, Jixuan Chen, Wenjing Hu, Xinyuan Wang, Yuhui Xu, Zekun Wang, Yiheng Xu, Junli Wang, Doyen Sahoo, Tao Yu, and Caiming Xiong. 2025.
\newblock \href {https://doi.org/10.48550/ARXIV.2505.13227} {{Scaling Computer-Use Grounding via User Interface Decomposition and Synthesis}}.
\newblock \emph{arXiv preprint}.

\bibitem[{Xie et~al.(2024)Xie, Zhang, Chen, Li, Zhao, Cao, Hua, Cheng, Shin, Lei, Liu, Xu, Zhou, Savarese, Xiong, Zhong, and Yu}]{osworld2024}
Tianbao Xie, Danyang Zhang, Jixuan Chen, Xiaochuan Li, Siheng Zhao, Ruisheng Cao, Toh~Jing Hua, Zhoujun Cheng, Dongchan Shin, Fangyu Lei, Yitao Liu, Yiheng Xu, Shuyan Zhou, Silvio Savarese, Caiming Xiong, Victor Zhong, and Tao Yu. 2024.
\newblock \href {https://arxiv.org/abs/2404.07972} {{OSWorld: Benchmarking Multimodal Agents for Open-Ended Tasks in Real Computer Environments}}.
\newblock \emph{Preprint}, arXiv:2404.07972.

\bibitem[{Xu et~al.(2024)Xu, Wang, Wang, Lu, Xie, Saha, Sahoo, Yu, and Xiong}]{aguvis2024}
Yiheng Xu, Zekun Wang, Junli Wang, Dunjie Lu, Tianbao Xie, Amrita Saha, Doyen Sahoo, Tao Yu, and Caiming Xiong. 2024.
\newblock \href {https://doi.org/10.48550/ARXIV.2412.04454} {{Aguvis: Unified Pure Vision Agents for Autonomous GUI Interaction}}.
\newblock \emph{arXiv preprint}.

\bibitem[{Yang et~al.(2023)Yang, Zhang, Li, Zou, Li, and Gao}]{yang2023setofmarkpromptingunleashesextraordinary}
Jianwei Yang, Hao Zhang, Feng Li, Xueyan Zou, Chunyuan Li, and Jianfeng Gao. 2023.
\newblock \href {https://arxiv.org/abs/2310.11441} {{Set-of-Mark} prompting unleashes extraordinary visual grounding in {GPT-4V}}.
\newblock \emph{Preprint}, arXiv:2310.11441.

\bibitem[{Yang et~al.(2025)Yang, Li, Dai, Yang, Luo, Zhao, Hu, Huang, Saha, Chen, Xu, Pan, Savarese, Xiong, and Li}]{gta12025}
Yan Yang, Dongxu Li, Yutong Dai, Yuhao Yang, Ziyang Luo, Zirui Zhao, Zhiyuan Hu, Junzhe Huang, Amrita Saha, Zeyuan Chen, Ran Xu, Liyuan Pan, Silvio Savarese, Caiming Xiong, and Junnan Li. 2025.
\newblock \href {https://doi.org/10.48550/ARXIV.2507.05791} {{GTA1: GUI Test-time Scaling Agent}}.
\newblock \emph{arXiv preprint}.

\bibitem[{Yao et~al.(2026)Yao, Xiong, Zou, Xiong, and Rong}]{autofocus2026}
Ruilin Yao, Shegnwu Xiong, Tianyu Zou, Shili Xiong, and Yi~Rong. 2026.
\newblock \href {https://doi.org/10.48550/ARXIV.2605.02630} {{AutoFocus: Uncertainty-Aware Active Visual Search for GUI Grounding}}.
\newblock \emph{arXiv preprint}.

\bibitem[{You et~al.(2024)You, Zhang, Schoop, Weers, Swearngin, Nichols, Yang, and Gan}]{ferretui2024}
Keen You, Haotian Zhang, Eldon Schoop, Floris Weers, Amanda Swearngin, Jeffrey Nichols, Yinfei Yang, and Zhe Gan. 2024.
\newblock \href {https://arxiv.org/abs/2404.05719} {{Ferret-UI: Grounded Mobile UI Understanding with Multimodal LLMs}}.
\newblock \emph{Preprint}, arXiv:2404.05719.

\bibitem[{Zhang et~al.(2025)Zhang, Xu, Zhu, Dai, Qiu, Yang, Luo, Chen, Wagle, Franklin, and Guo}]{phiground2025}
Miaosen Zhang, Ziqiang Xu, Jialiang Zhu, Qi~Dai, Kai Qiu, Yifan Yang, Chong Luo, Tianyi Chen, Justin Wagle, Tim Franklin, and Baining Guo. 2025.
\newblock \href {https://doi.org/10.48550/ARXIV.2507.23779} {{Phi-Ground Tech Report: Advancing Perception in GUI Grounding}}.
\newblock \emph{arXiv preprint}.

\end{thebibliography}
\appendix
\section*{Appendix}
\section{Prompt Templates}
\label{sec:prompts}

We list the task-specific prompts used by \method.
Braced fields, highlighted in dark red, denote example-specific placeholders.
Image placeholders represent image attachments rather than literal prompt text.
Both auxiliary tasks share the system message shown below and include one image in the user message.
The grounder retains its model-specific chat template and coordinate-output instructions.

\paragraph{Center prediction.}
The image is the unmarked screenshot during training and the current unmarked view during inference.
The same prompt is used for both center-prediction stages.
\begin{promptbox}
\textbf{[System]}\\
You are a helpful assistant.

\textbf{[User]}\\
\promptfield{\{unmarked\_image\}}\\
Locate the UI element requested by the instruction.\\
Instruction: \promptfield{\{instruction\}}

Return only <center>x, y</center>.\\
Coordinates are normalized to 0--1000 within this image.
\end{promptbox}

\paragraph{Candidate verification and description.}
The input image is the enlarged view with a translucent red circle at the candidate center.
Candidate coordinates are normalized within this view.
\begin{promptbox}
\textbf{[System]}\\
You are a helpful assistant.

\textbf{[User]}\\
\promptfield{\{marked\_view\}}\\
Instruction: \promptfield{\{instruction\}}\\
Candidate center: (\promptfield{\{x\}}, \promptfield{\{y\}})\\
Coordinates are normalized to 0--1000 within this image.\\
The translucent red circle marks the candidate center.

If this candidate is the requested target, return <desc>description</desc>.\\
Otherwise, return <desc>REJECT</desc>.
\end{promptbox} 

\paragraph{Grounding with paired cues.}
For an accepted candidate that passes consistency filtering, the grounding model sees the same marked view.
The following augmented instruction is embedded in its native grounding prompt:
\begin{promptbox}
\textbf{[System]}\\
...

\textbf{[User]}\\
\promptfield{\{marked\_view\}}\\
...\\
\promptfield{\{instruction\}}

The red circle is a location hint. If you are unsure where to click, click the center of the red circle.\\
The marked element is described as: \promptfield{\{description\}}\\
...
\end{promptbox}
Ellipses denote unchanged model-specific system messages, tool definitions, and user-prompt wrappers. The system message is omitted when absent from the native template.
The first two appended sentences form the fixed location hint and belong to the Visual Cue.
The final sentence introduces the generated Textual Cue.
When verification rejects the candidate but consistency filtering passes, only the marker and location hint are retained.
When consistency filtering fails, View-only fallback uses the unmarked view and original instruction without either cue.
\section{Evaluation Details}
\label{sec:implementation}

\paragraph{Benchmark statistics.}
Table~\ref{tab:benchmark-statistics} summarizes the evaluation benchmark sizes.
OSW-G Original and Refined each contain $564$ tasks and share screenshots.
MMB-GUI contains $1{,}787$ Basic and $1{,}807$ Advanced examples.

\begin{table}[!htbp]
    \centering
    \small
    \begin{tabular}{lr}
    \toprule
    \textbf{Benchmark} & \textbf{Samples} \\
    \midrule
    SSPro & $1{,}581$ \\
    OSW-G$_{\mathrm{R}}$ & $564$ \\
    MMB-GUI & $3{,}594$ \\
    SSv2 & $1{,}272$ \\
    UI-V & $5{,}479$ \\
    \bottomrule
    \end{tabular}
    \caption{Sample counts for the evaluation benchmarks.}
    \label{tab:benchmark-statistics}
\end{table}

\paragraph{Visual budget settings.}
We compare $12.5\%$-area crops enlarged at linear scales $1$, $\sqrt{2}$, and $2$.
We also evaluate $25\%$-area crops at scales $1$ and $\sqrt{2}$.
Each configuration uses the full grounding-evaluation sets for OSW-G$_{\mathrm{R}}$, SSPro, and UI-V listed in Table~\ref{tab:benchmark-statistics}.
\section{Qualitative Case Studies}
\label{sec:case-studies}

\begin{figure*}[t]
\centering
\definecolor{casegt}{HTML}{16845B}
\definecolor{casecrop}{HTML}{B77926}
\definecolor{casecorrect}{HTML}{0072B2}
\definecolor{casewrong}{HTML}{A63860}
\definecolor{casemarker}{HTML}{C52222}
\footnotesize
{\textcolor{casegt}{$\square$} Target (GT)\quad
 \textcolor{casecrop}{- -} Crop boundary\quad
 \textcolor{casewrong}{$\times$} Wrong click\quad
 \textcolor{casecorrect}{$\bullet$} Correct click\quad
 \textcolor{casemarker}{$\circ$} Input marker\par}
\newcommand{\previewcase}[7]{%
    \begin{tcolorbox}[colback=black!3,colframe=black!18,boxrule=0.4pt,arc=1mm,
        left=5pt,right=5pt,top=5pt,bottom=5pt,before skip=6pt,after skip=3pt,
        fontupper=\footnotesize]
    \raggedright
    {\raggedright\noindent\textbf{Instruction:} #2\par}
    \vspace{3pt}
    \noindent\begin{minipage}[b]{.205\linewidth}\centering\scriptsize #3\end{minipage}\hfill
    \begin{minipage}[b]{.365\linewidth}\centering #4\end{minipage}\hfill
    \begin{minipage}[b]{.365\linewidth}\centering #5\end{minipage}\par
    \includegraphics[width=\linewidth]{#6}\par
    \if\relax\detokenize{#7}\relax\else
    \vspace{2pt}
    \noindent Cue: #7\par
    \fi
    \end{tcolorbox}
    {\centering\textbf{#1}\par}
}

\previewcase
    {(a) Corrected with View}
    {Cut}
    {SSPro\\$2560\times1440$ pixels}
    {\textcolor{casewrong}{Full: $\times$}}
    {\textcolor{casecorrect}{w. View: $\checkmark$}}
    {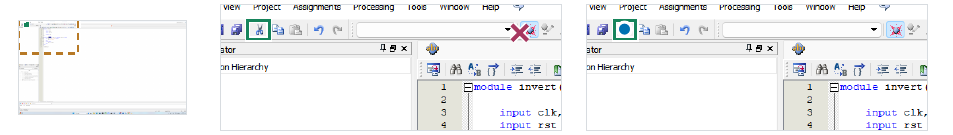}
    {The Cut icon is located in the toolbar, represented by a pair of scissors.}

\previewcase
    {(b) Corrected with View \& Cue}
    {view more option of edit button}
    {SSv2\\$2560\times1440$ pixels}
    {\textcolor{casewrong}{w. View: $\times$}}
    {\textcolor{casecorrect}{w. View \& Cue: $\checkmark$}}
    {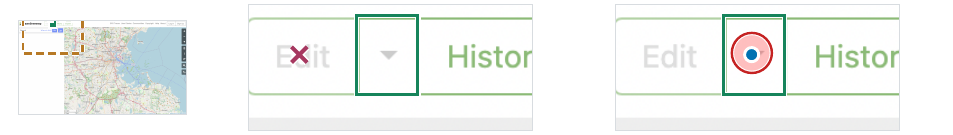}
    {It is a small downward arrow that allows users to expand additional options related to editing.}

\previewcase
    {(c) Misleading cue}
    {check the date and time menu}
    {SSv2\\$960\times540$ pixels}
    {\textcolor{casecorrect}{w. View: $\checkmark$}}
    {\textcolor{casewrong}{w. View \& Cue: $\times$}}
    {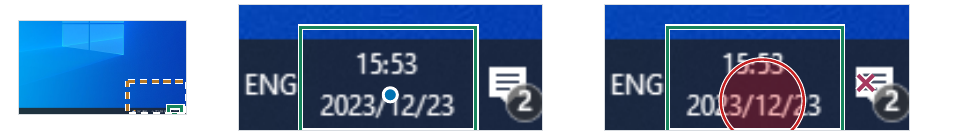}
    {It is represented by an icon with a speech bubble and the number `2' inside it.}

\previewcase
    {(d) Target excluded by cropping}
    {Click on the U42 (21st column from the left, 42nd row from the top) in the sheet}
    {OSW-G$_{\mathrm{R}}$\\$1920\times1080$ pixels}
    {\textcolor{casecorrect}{Full: $\checkmark$}\\\scriptsize Target region (outside the view)}
    {\textcolor{casewrong}{w. View: $\times$}\\\scriptsize Selected local region}
    {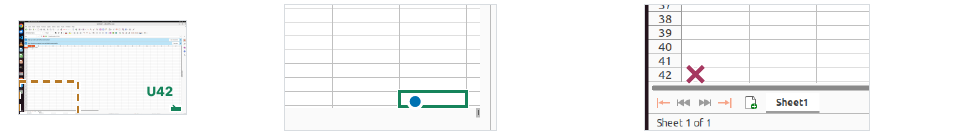}
    {}

\caption{Successful corrections and failure cases of view and cue preprocessing for Qwen3-VL-2B.}
\label{fig:case-studies}
\end{figure*}

Figure~\ref{fig:case-studies} compares the same frozen Qwen3-VL-2B-Instruct model using the full screenshot (Full), the preprocessed view (w. View), and that view with paired cues (w. View \& Cue).
Both preprocessed-view conditions use the same predicted center, $12.5\%$ cropping area, and $2\times$ linear enlargement.
All four cases have accepted descriptions, so rejection handling does not affect these comparisons.
The examples are selected to illustrate distinct outcomes, not to estimate their frequency.
They show the inputs before consistency filtering.
Detail panels use the same region in (a--c).
Panel (d) contrasts the excluded target with the selected region.
Cue text is excerpted verbatim.
Boxes, click symbols, and marker outlines are added for clarity.
The translucent red disk is a model input.

Section~\ref{sec:cue-reliability} analyzes how these cases distinguish visibility, target interpretation, and information loss.
\section{Detailed Benchmark Results}
\label{sec:benchmark-breakdown}

Tables~\ref{tab:breakdown-uivision}--\ref{tab:breakdown-mmbench} report Qwen3-VL-2B and Qwen3-VL-4B results with Full inputs and with \method{}.
Full denotes our reproduced baseline using the original screenshot.
All scores are accuracies (\%), and $N$ is the number of examples in each category.
We follow the benchmark groupings used by \citet{groundcua2025}, with additional UI-V element-type results.
The \method{} configuration is identical to the main experiments, including the $2\%$ center-consistency filter and unmarked-view fallback.
These are breakdowns of the same evaluated samples, rather than additional independent evaluations.

\paragraph{Aggregation.}
SSPro uses application domains crossed with Text/Icon labels.
MMB-GUI uses platforms crossed with Basic/Advanced difficulty.
UI-V instruction categories are Basic, Functional, and Spatial.
Its Text/Icon rows provide a separate partition of the same examples.
Overall accuracy is computed over examples, except for MMB-GUI, where it is the mean of Basic and Advanced accuracies.

\paragraph{OSW-G categories.}
We report refined instructions only.
GUI task labels overlap, so their counts cannot be added to obtain the overall sample count.
The released GroundCUA category-matching code assigns nine examples to its \emph{Refusal bucket}, whereas the benchmark contains 54 refusal tasks according to its target-presence annotations.
Table~\ref{tab:breakdown-osworld} distinguishes these two definitions and additionally reports the 510 target-present examples.
The overall score includes all 564 examples.

\begin{table}[!htbp]
\centering
\footnotesize
\setlength{\tabcolsep}{3pt}
\begin{tabular*}{\columnwidth}{@{\extracolsep{\fill}}lrcccc@{}}
\toprule
 & & \multicolumn{2}{c}{\textbf{Qwen3-VL-2B}} & \multicolumn{2}{c}{\textbf{Qwen3-VL-4B}} \\
\cmidrule(lr){3-4}\cmidrule(l){5-6}
\textbf{Category} & $N$ & Full & \method{} & Full & \method{} \\
\midrule
Basic & $1772$ & $23.8$ & $43.8$ & $33.5$ & $49.6$ \\
Functional & $1772$ & $21.5$ & $41.8$ & $31.5$ & $47.5$ \\
Spatial & $1935$ & $5.3$ & $17.0$ & $12.0$ & $26.0$ \\
Text & $980$ & $42.6$ & $61.2$ & $54.1$ & $69.7$ \\
Icon & $4499$ & $10.9$ & $27.7$ & $19.0$ & $34.2$ \\
\midrule
Overall & $5479$ & $16.5$ & $33.7$ & $25.3$ & $40.6$ \\
\bottomrule
\end{tabular*}
\caption{UI-V results by instruction and element type.}
\label{tab:breakdown-uivision}
\end{table}
\begin{table}[!htbp]
\centering
\footnotesize
\setlength{\tabcolsep}{3pt}
\begin{tabular*}{\columnwidth}{@{\extracolsep{\fill}}lrcccc@{}}
\toprule
 & & \multicolumn{2}{c}{\textbf{Qwen3-VL-2B}} & \multicolumn{2}{c}{\textbf{Qwen3-VL-4B}} \\
\cmidrule(lr){3-4}\cmidrule(l){5-6}
\textbf{Category} & $N$ & Full & \method{} & Full & \method{} \\
\midrule
CAD/Text & $197$ & $32.0$ & $61.4$ & $59.9$ & $68.0$ \\
CAD/Icon & $64$ & $18.8$ & $28.1$ & $20.3$ & $32.8$ \\
Dev/Text & $154$ & $64.3$ & $74.0$ & $81.8$ & $78.6$ \\
Dev/Icon & $145$ & $13.1$ & $37.2$ & $34.5$ & $40.0$ \\
Creative/Text & $198$ & $57.6$ & $67.7$ & $71.2$ & $71.7$ \\
Creative/Icon & $143$ & $18.2$ & $30.8$ & $25.2$ & $33.6$ \\
Scientific/Text & $144$ & $61.8$ & $75.0$ & $81.2$ & $84.7$ \\
Scientific/Icon & $110$ & $22.7$ & $40.9$ & $33.6$ & $43.6$ \\
Office/Text & $177$ & $72.9$ & $85.3$ & $87.6$ & $88.1$ \\
Office/Icon & $53$ & $37.7$ & $58.5$ & $45.3$ & $60.4$ \\
OS/Text & $107$ & $59.8$ & $69.2$ & $78.5$ & $70.1$ \\
OS/Icon & $89$ & $24.7$ & $49.4$ & $40.4$ & $56.2$ \\
All/Text & $977$ & $57.1$ & $71.9$ & $75.8$ & $76.8$ \\
All/Icon & $604$ & $20.5$ & $39.1$ & $32.5$ & $42.5$ \\
\midrule
Overall & $1581$ & $43.1$ & $59.3$ & $59.3$ & $63.7$ \\
\bottomrule
\end{tabular*}
\caption{SSPro results by application domain and element type.}
\label{tab:breakdown-sspro}
\end{table}
\begin{table}[!htbp]
\centering
\footnotesize
\setlength{\tabcolsep}{3pt}
\begin{tabular*}{\columnwidth}{@{\extracolsep{\fill}}lrcccc@{}}
\toprule
 & & \multicolumn{2}{c}{\textbf{Qwen3-VL-2B}} & \multicolumn{2}{c}{\textbf{Qwen3-VL-4B}} \\
\cmidrule(lr){3-4}\cmidrule(l){5-6}
\textbf{Category} & $N$ & Full & \method{} & Full & \method{} \\
\midrule
Text Matching & $261$ & $74.7$ & $73.2$ & $81.2$ & $78.5$ \\
\shortstack[l]{Element\\recognition} & $330$ & $66.1$ & $72.7$ & $73.3$ & $75.5$ \\
\shortstack[l]{Layout\\understanding} & $253$ & $71.9$ & $74.3$ & $76.7$ & $79.4$ \\
\shortstack[l]{Fine-grained\\manipulation} & $152$ & $48.7$ & $52.0$ & $62.5$ & $59.9$ \\
Refusal bucket & $9$ & $0.0$ & $0.0$ & $22.2$ & $0.0$ \\
All refusal tasks & $54$ & $0.0$ & $0.0$ & $11.1$ & $3.7$ \\
Target-present & $510$ & $68.6$ & $73.9$ & $76.9$ & $77.5$ \\
\midrule
Overall & $564$ & $62.1$ & $66.8$ & $70.6$ & $70.4$ \\
\bottomrule
\end{tabular*}
\caption{OSW-G$_{\mathrm{R}}$ results by task type.}
\label{tab:breakdown-osworld}
\end{table}
\begin{table}[!htbp]
\centering
\footnotesize
\setlength{\tabcolsep}{3pt}
\begin{tabular*}{\columnwidth}{@{\extracolsep{\fill}}lrcccc@{}}
\toprule
 & & \multicolumn{2}{c}{\textbf{Qwen3-VL-2B}} & \multicolumn{2}{c}{\textbf{Qwen3-VL-4B}} \\
\cmidrule(lr){3-4}\cmidrule(l){5-6}
\textbf{Category} & $N$ & Full & \method{} & Full & \method{} \\
\midrule
Windows/Basic & $271$ & $80.4$ & $87.8$ & $86.3$ & $90.4$ \\
Windows/Adv & $272$ & $45.6$ & $52.6$ & $68.0$ & $59.9$ \\
macOS/Basic & $345$ & $78.0$ & $87.8$ & $90.4$ & $89.9$ \\
macOS/Adv & $346$ & $50.3$ & $61.6$ & $73.1$ & $68.5$ \\
Linux/Basic & $191$ & $66.0$ & $75.4$ & $79.6$ & $77.0$ \\
Linux/Adv & $196$ & $45.4$ & $58.2$ & $59.7$ & $60.2$ \\
iOS/Basic & $314$ & $94.3$ & $90.1$ & $96.8$ & $93.3$ \\
iOS/Adv & $330$ & $76.4$ & $73.9$ & $86.1$ & $75.8$ \\
Android/Basic & $356$ & $90.7$ & $91.9$ & $95.8$ & $91.9$ \\
Android/Adv & $355$ & $77.7$ & $72.1$ & $88.7$ & $75.2$ \\
Web/Basic & $310$ & $88.4$ & $92.3$ & $97.4$ & $94.2$ \\
Web/Adv & $308$ & $63.6$ & $69.2$ & $79.5$ & $69.5$ \\
All/Basic & $1787$ & $84.3$ & $88.5$ & $92.1$ & $90.3$ \\
All/Adv & $1807$ & $61.5$ & $65.5$ & $77.4$ & $69.1$ \\
\midrule
Overall & $3594$ & $72.9$ & $77.0$ & $84.7$ & $79.7$ \\
\bottomrule
\end{tabular*}
\caption{MMB-GUI results by platform and difficulty.}
\label{tab:breakdown-mmbench}
\end{table}
\FloatBarrier
\clearpage
\clearpage
\end{document}